\documentclass[journal]{IEEEtran} 
\usepackage[english]{babel}
\usepackage{amssymb}

\usepackage{mathtools, amsmath, amssymb,bbm}
\usepackage{cite}

\ifCLASSINFOpdf
  \usepackage[pdftex]{graphicx}
\else
\fi 

\usepackage{amsmath}
\usepackage{algorithm}
\usepackage{algpseudocode}

\usepackage{tikz,xcolor}

\definecolor{lime}{HTML}{A6CE39}
\DeclareRobustCommand{\orcidicon}{%
	\begin{tikzpicture}
	\draw[lime, fill=lime] (0,0) 
	circle [radius=0.16] 
	node[white] {{\fontfamily{qag}\selectfont \tiny ID}};
	\draw[white, fill=white] (-0.0625,0.095) 
	circle [radius=0.007];
	\end{tikzpicture}
	\hspace{-2mm}
}
\foreach \x in {A, ..., Z}{%
	\expandafter\xdef\csname orcid\x\endcsname{\noexpand\href{https://orcid.org/\csname orcidauthor\x\endcsname}{\noexpand\orcidicon}}
}

\begin{document}
\title{Particle Swarm Optimization for Energy Disaggregation in Industrial and Commercial Buildings}

\author{Karoline~Brucke\orcidA{}, Stefan~Arens\orcidB{}, Jan-Simon~Telle\orcidC{}, Sunke~Schlüters\orcidD{}, Benedikt~Hanke\orcidE{}, Karsten~von~Maydell\orcidF{} and Carsten~Agert\orcidG{}

\thanks{Corresponding author: Jan-Simon Telle, email: jan-simon.telle@dlr.de}
\thanks{All authors were with DLR-Institute of Networked Energy Systems, Carl-von-Ossietzky-Str. 15, 26129 Oldenburg, Germany}
\thanks{Manuscript uploaded to arxiv.org on \today}
}



\maketitle

\begin{abstract}
This paper provides a formalization of the energy disaggregation problem for particle swarm optimization and shows the successful application of particle swarm optimization for disaggregation in a multi-tenant commercial building. The developed mathmatical description of the disaggregation problem using a state changes matrix belongs to the group of non-event based methods for energy disaggregation. This work includes the development of an objective function in the power domain and the description of position and velocity of each particle in a high dimensional state space. For the particle swarm optimization, four adaptions have been applied to improve the results of disaggregation, increase the robustness of the optimizer regarding local optima and reduce the computational time. The adaptions are varying movement constants, shaking of particles, framing and an early stopping criterion. In this work we use two unlabelled power datasets with a granularity of 1\,s. Therefore, the results are validated in the power domain in which good results regarding multiple error measures like root mean squared error or the percentage energy error can be shown. 
\end{abstract}


\IEEEpeerreviewmaketitle

\section{Introduction}
\begin{figure}[!t]
\centering
\includegraphics[width=3.5in]{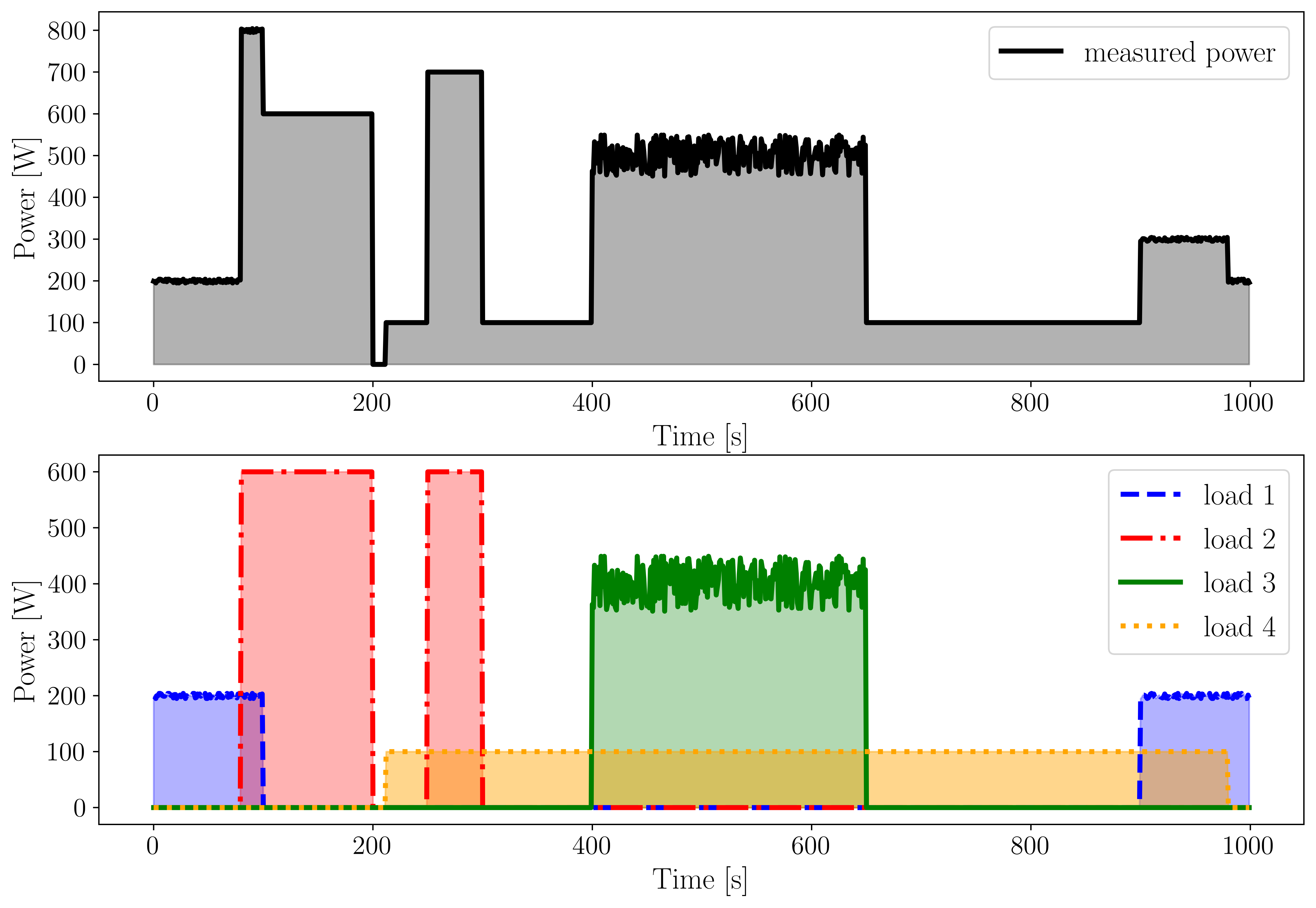}
\caption{Graphical representation of energy disaggregation. Upper illustration shows the total measured power. The bottom illustration shows the corresponding power of four individual loads over time, totaling to be the measured aggregate power.}
\label{fig:NILM}
\end{figure}

\IEEEPARstart{D}{ue} to the increasing share of renewable energies in the electricity generation, the electricity supply is getting more volatile. In order to guarantee stability of the power grid, adaptions on both the producer and the consumer side are getting more important \cite{kirschen2003}. Adaptions on the consumer side are called demand side management (DSM) \cite{palensky2011,strbac2008}. DSM in buildings is carried out by energy management systems. Energy management can be realized by submetering as in the often used REDD dataset \cite{kolter2011redd}. But that leads to great amounts of data and is hardly feasible for a large number of buildings. In order to reduce the needed data, non-intrusive load monitoring (NILM) can be used which has been described first by Hart in \cite{Hart1992}. The objective of NILM is the description of the state of every device in an aggregate power signal without a complex submetering \cite{Goncalves2011,Zeifman2012}. Figure~\ref{fig:NILM} shows the principle of NILM where a measured aggregate power signal is divided into the single contributions of individual loads. Thus, NILM is also called energy disaggregation.  

Multiple publications are working on this topic with event based methods \cite{Liang2010} and non-event based methods \cite{cominola2017} and methods from supervised and unsupervised machine learning \cite{Pattem2012,Kim2011,Parson2014,Salerno2018}. However, in this context almost exclusively households are considered, since the energy systems of individual households are of manageable complexity \cite{Kelly2015}. Industrial and commercial buildings are hardly investigated due to their complex energy systems and often confidential electricty data \cite{Kriechbaumer2018}. But due to the high energy demand of industrial buildings, there are large saving potentials and great possibilities for effective demand-side management. Since commercial properties in particular, can differ greatly from one another, approaches and algorithms are needed that work independently of these differences and adapt to any dataset. Additionally, often there is no other knowledge than the measured data. Thus, machine learning algorithm requiring a lot of prior knowledge and many features are not feasible for NILM approaches for indutrial buildings. Furthermore, machine learning methods often rely on a costly training regarding computational time and power for operation and amount of data.

In this work, we present a new, fully unsupervised approach using particle swarm optimization (PSO) for energy disaggregation and show its application in a multi-tenant commercial building. The disaggregation problem is of high complexity. PSO is able to find complex solutions even if the information for each particle is limited \cite{Dreo2006}. It has been implemented for various applications and complex real world problems \cite{shi2001, del2008}. The method has been very successful in solving high dimensional nonlinear optimization problems also in applications of power system \cite{miranda2002, alrashidi2006,alrashidi2008}. In theory, PSO has been deployed effectively to the multidimensional \textit{Knapsack} problem \cite{chu1998}. It is very similar to the formulation of the disaggregation problem stated in this work. However, applications to real world disaggregation problems left room for improvements in the past due to a lack of descriptive characteristics of the single devices \cite{egarter2015}. On the other hand, using the metaheuristic PSO no complex training or model building is necessary. The method does not adapt to the data during optimization. That could increase the transferability to other datasets with minimal changes. Aditionally, problems like underfitting and overfitting due to the complexity of chosen models do not occur. 

For developing and testing the method, three phase power data in active and reactive power of a multi tenant commercial building and the according device profiles was used. This measured power data is referred to as aggregate power signal. The device profiles must not be full appliance signatures but e.g. one component of an appliance signature like one operational mode of a complex device. The developed method takes the measured aggregate power signal and device profiles as input to determine the state of each device for any given point in time.  

In the first part of this paper we present our formulation of the disaggregation problem and how PSO can be used and improved for energy disaggregation. Therefore, we firstly introduce the classic PSO method followed by the formal description of the state space, position and velocity of the PSO for energy disaggregation. Thereafter, the adaptions to PSO for energy disaggregation are stated. The adaptions are time varying movement constants, shaking of particles, framing of the power signal and an early stopping criterion. To our knowledge, this combination of adaptions to PSO is new. In the second part of this work, the testing of the developed method is described. Therein, the used data and error measures are presented. The results and their discussion are outlined subsequently. The paper is closed with a conclusion and outlook.

\section{Particle Swarm Optimization (PSO) For Energy Disaggregation}

In this work, we present the application of PSO to the field of NILM  as a posibility of a fully unsupervised disaggregation. The assumed standard granularity of the power data is of 1\,s. Let $T$ denote the time domain, i.e. in this case $T = \{ 0, 1, \dotsc, 86399\}$ for the duration of one day. We assume that the power $P$ consists of six features, three phases of active power and three phases of reactive power. We think of the power $P$ as a function mapping the timestep to the six power values, i.e.: \begin{equation}
    P : T \to \mathbb{R}^6
\end{equation}

\begin{figure}[!t]
\centering
\includegraphics[width=3.5in]{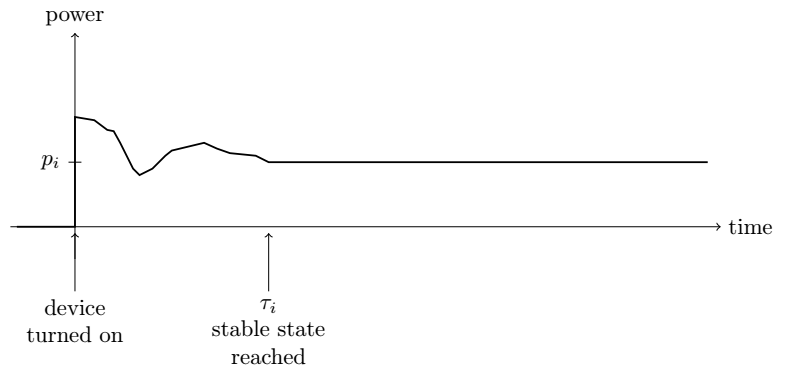}
\caption{Theoretical representation of a device profile $l_{i}$. Until $\tau_{i}$ the device shows a dynamic behavior after that it is operating in a stable state with a constant power consumption of $p_{i}$.}
\label{fig:device_profile} 
\end{figure}

We assume, that the device profile of device $i$ is characterized by a transient or dynamic profile after the device is switched on and a constant profile after the device reached a stable state. This behaviour is shown in Figure~\ref{fig:device_profile}. We denote the device profile by:

\begin{equation}
    l_i : \mathbb{Z} \to \mathbb{R}^6 \,.
\end{equation}

Let $p_{i} \in \mathbb{R}^6$ denote the power consumption in the stable operating state and $\tau_{i}$ the (typical) time until this state is reached. By convention we set $l_{i}(t) = p_{i}$ for $t>\tau_{i}$ regardless of the actual runtime of the device. Figure~\ref{fig:device_profile} is a graphical representation of a device profile according to the assumptions made. 

We assume the aggregated  power $P(t)$ can be described at any time $t$ as a superposition of the power of all individual device types $1 \dots M$. By $S \in \{0,1,-1\}^{T \times M}$ we denote the state change matrix.  The entry at the $t$\textsuperscript{th} row and the $i$\textsuperscript{th} column denotes a possible state change of device $i$ at time $t$ ($1$ meaning the device was switched on, $-1$ the device was switched off and 0 that the state remains the same). It will be convienient to denote the $i$\textsuperscript{th} column by $s_{i}$ and interpret it as a function of time:

\begin{equation}
s_{i}: T \to \{0,1,-1\}
\end{equation}
with $s_{i}(t)$ denoting the entry at the i-th column and t-th row of $S$. Entries in $S$ unequal zero are referred to as \textit{events} in the following, with $s_{i}(t) = 1$ being \textit{ON-events} and $s_{i}(t) = -1$ being \textit{OFF-events} respectively.  The aggregate power signal can then be described by:

\begin{multline}
    P(t) = \sum_{\substack{i,\tilde t\\s_{i}(\tilde t) = 1}} s_{i}(\tilde t) l_{i}(t + \tilde t) +\\
     \sum_{\substack{i,\tilde t\\s_{i}(\tilde t) = -1}} s_{i}(\tilde t) \mathbbm{1}_{(\tilde t,T)}(t) p_{i} + \epsilon(t)
\label{eq:aggregate}
\end{multline}
where $\epsilon(t)$ is referred to as always-on-component or noise. The first summand in Equation~\ref{eq:aggregate} describes all ON-events. If $s_{i}(t) = 1$, the total device profile including dynamic behavior and stable state is added to the aggregate power. The second summand in Equation~\ref{eq:aggregate} describes the OFF-events resprectively. If $s_{i}(t) = -1$, only the power value of the device's stable state is substracted. Given these assumptions for the aggregate power signal, the following optimization problem has to be solved: 

\begin{equation}
    \min_S E \bigg( P, P_{\mathrm{S}}  \bigg)
\label{eq:problem}
\end{equation}
where
\begin{multline}
    P_{\mathrm{S}} (t) = \sum_{i} \sum_{\tilde t} s_{i}(\tilde t) l_i(t + \tilde t)+\\ \sum_{\substack{i,\tilde t\\s_i(\tilde t) = -1}} s_i(\tilde t) \mathbbm{1}_{(\tilde t,T)}(t) p_i
\label{eq:reconstruction}
\end{multline}
and $E(P,P_{\mathrm{S}})$ being an error function of $P$ and $P_{\mathrm{S}}$. $P$ denotes the measured aggregate power signal. $P_{\mathrm{S}}$ denotes the reconstructed or approximated power according to Equation~\ref{eq:reconstruction}, using the state changes matrix $S$ and the device profiles $l_i$. The state changes matrix $S$ is optimized by PSO, with $E(P, P_{\mathrm{S}})$ being the objective function of the optimizer. For a high number of timesteps and device profiles, the complexity of the disaggregation problem increases rapidly. For this reason, we appled four adaptions of the PSO for a faster and more robust disaggregation of the measured power data. Thus, in the following sections the classic PSO method is stated first. Thereafter, the adaptions to the PSO are presented. Finally, the whole algorithm will be stated in pseudocode. 

\subsection{Classic PSO}
PSO was developed by the authors Kennedy and Eberhart in 1995 as an analogy to the collective behavior of movement of and within large groups of animals \cite{Kennedy1995,Kennedy1995_2}. Even if the information available to individuals in swarms is limited, complex movement patterns can be observed \cite{Dreo2006}. Formally, the PSO consists of a swarm of $N$ particles each moving with a certain velocity $\vec{v}_{n}(i)$ through the (multidimensional) state space. Each position $\vec{v}_{n}(i)$ in the state space is a possible solution of the optimization problem and assigned a value of an objective function, which has to be optimized. In every iteration, a position update is calculated for each particle according to:

\begin{equation}
\label{eq:pos_update}
\vec{x}_{n}(i+1) = \vec{x}_{n}(i) + \vec{v}_{n}(i)
\end{equation}
The velocity is dependent on two quantities: The best past position of the particle $\vec{x}_{\mathrm{n,best}}$ and the best global position $\vec{x}_{\mathrm{g,best}}$ \cite{Dreo2006,Kennedy1995}. 
 
\begin{multline}
\vec{v}_{n}(i+1) = a_{\mathrm{t}} \cdot \vec{v}_{n}(i) + \phi_1 (\vec{x}_{\mathrm{n,best}} - \vec{x}_{n}(i))\\ + \phi_2 (\vec{x}_{\mathrm{g,best}} - \vec{x}_{n}(i))
\label{eq:vel_update}
\end{multline}

The constant $a_{\mathrm{t}}$ describes an inertia with respect to the previous velocity. $\phi_1$ and $\phi_2$ contain a random part described by the random constants $r_1$ and $r_2$ and a not random part described by the parameters $a_{\mathrm{k}}$ and $a_{\mathrm{s}}$. $a_{\mathrm{k}}$ and $a_{\mathrm{s}}$ are called \textit{cognitive} and \textit{social} constant.

\begin{equation}
\phi_1 = a_{\mathrm{k}} \cdot r_1 \text{ and } \phi_2 = a_{\mathrm{s}} \cdot r_2
\end{equation} 

The constants $a_{\mathrm{t}}$, $a_{\mathrm{k}}$ and $a_{\mathrm{s}}$ have to be adapted to the problem for better and faster convergence in the optimization process of the PSO as \cite{trelea2003,melin2013,alireza2011} show. The ratio of $a_{\mathrm{s}}$ and $a_{\mathrm{k}}$ set the level of importance given the particles movement either its best previous position or the current best global position~\cite{melin2013}. Therefore, the ratio determines the direction in which the velocity-vector is pointing. 

Core of the PSO method is its objective function which guides the search. It has to be defined beforehand and for the proposed definition of the PSO the objective function has to result in a scalar.   

\subsection{Formalization of the Disaggregation Problem for PSO}

Generally, the disaggregation problem, as presented in Equation~\ref{eq:problem} with respect to Equation~\ref{eq:reconstruction}, searches for the state changes matrix $S$ that minimizes the error function $E$. $S$ is optimized by the PSO. The selected power data for the disaggregation is restricted to the length of one day. Thus, every day in the used power datasets is disaggregated in one run of the algorithm to limit the propagation of a possible error. In general, the split of the day needs to be at the time with the lowest probability for any device to be running. Thus, the time of splitting the individual days is at midnight. In order to use particle swarm optimization for disaggregation of the measured aggregate power $P$, the particle swarm optimizer is adapted to the problem. This includes firstly the description of the state space and secondly the choice of a suitable objective function. 

The state space is described by $S$ from Equation~\ref{eq:reconstruction}. For disaggregation, the PSO trys to find the state $s_{i}(t)$ of each individual device for each point in time $t$ in order to minimize the respective error function. Therefore, we denote the position of particle $p$ as a state changes matrix $S_{p} \in \mathbb{R}^{T\times M}$. We use no constraining function to limit the entries of $S_{p}$. Therefore, $S_{p}$ can contain non-discrete values. Before the evaluation of fitness of the respective particle we discretise $S_{p}$ in order to be able to apply Equation~\ref{eq:reconstruction}. Therefore, a threshold $r_{\mathrm{u}}$ is defined which defines an element in the state changes matrix to be an event. In this case, the threshold is chosen to be $r_{\mathrm{u}} = 0.6$. Thus, if the absolute value of an entry in $S_{p}$ is greater than $r_{\mathrm{u}}$ it gets an event. With this definition of an event-trheshold it is possible that little changes in the position matrix of particles, particularly around zero, do not influence the particle's fitness i.e. it's respective error value. 

Since the correct state change matrix is not known, but only the measured total power $P$, the objective function for the PSO must be calculated in the power domain for every feature measured (e.g. active and reactive power). For this purpose, the power is reconstructed in each iteration for each particle according to Equation~\ref{eq:reconstruction} from the current position of the particle. The reconstructed power according to the particle swarm optimization of $S$ is referred to as $P_{\mathrm{S}}$ in the following.  For the description of the objective function an error measure $E$ is developed, which results from two contributions. The first summand in Equation~\ref{eq:error_disaggregation} describes a squared deviation of the measured signal relative to the reconstructed signal. The second summand in Equation~\ref{eq:error_disaggregation} represents a squared deviation in the derivative of both quantities. The derivative is taken into account for being able to penalize deviations in sharp rises or peaks more heavily. $E$ is calculated for a specific interval $[a,b)$ and thus describes the deviation between measured and approximated power on this interval.

\begin{multline}
E^{[a,b)}(P,P_\mathrm{S}) = \alpha \cdot \sum_{t=a}^{b-1} (\vec{P}_{\mathrm{S}}(t)-\vec{P}(t))^2 +\\
 \beta \cdot \sum_{t=a}^{b-2} (\Delta \vec{P}_{\mathrm{S}}(t)-\Delta \vec{P}(t))^2
\label{eq:error_disaggregation}
\end{multline}

The parameters $\alpha$ and $\beta$ obey the condition $\alpha + \beta = 1$ and set the relation between the summands to each other. In this paper, $\alpha = 0.9$ and $\beta = 0.1$ are chosen empirically but the relation of $\alpha$ and $\beta$ can be adapted to the present power data as suitable. The calculation of the time derivative of the measured and reconstructed power is done according to the following Equation~\ref{eq:derivative} whereas $t+1$ denotes the subsequently measured timestep with respect to $t$. 

\begin{equation}
 \Delta \vec{P}(t) = \frac{\vec{P}(t+1)-\vec{P}(t)}{(t+1)-t} 
 \label{eq:derivative}  
\end{equation}

The proposed objective function describes an error measure and has to be minimized during optimization as in Equation~\ref{eq:problem}.

\subsection{Improvements of PSO for Disaggregation}
The results and the optimization process of the PSO are improved by adapting the method to the problem. These adjustments are described in the following subchapters.\\

\subsubsection{Time varying constants}

The velocity parameters $a_{\mathrm{s}}$ and $a_{\mathrm{k}}$ are responsible for the type of movement of the particles. If $a_{\mathrm{k}} > a_{\mathrm{s}}$ is selected, the particles move more sporadically and explorative through the state space  \cite{Kennedy1995}. This means that a large part of the state space is seen and evaluated by the particles. However, it also carries the risk that the probability of converging into the global minimum with smaller step sizes decreases. If $a_{\mathrm{s}} > a_{\mathrm{k}}$ is chosen, the swarm of particles moves more closed through the state space, but this can lead to a convergence of the whole swarm in an early discovered local optimum. Therefore, the adjustment of $a_{\mathrm{s}}$ and $a_{\mathrm{k}}$ describes the trade off between exploration and exploitation. To overcome the described difficulties, time varying constants have been proposed in the past by e.g. \cite{Ratnaweera2004} and have often been applied and further improved \cite{cai2009,mohammadi2012}. This procedure is also similar to the gradual change of the learning rate when using neural networks \cite{Battiti1989}. In this work, we use linear changing velocity parameters. At the beginning of the optimization, $a_{\mathrm{k}} >> a_{\mathrm{s}}$ should apply in order to evaluate as much of the state space as possible. During optimization,  $a_{\mathrm{k}}$ then decreases and $a_{\mathrm{s}}$ increases. Specifically, the linear change of the constants over $I$ iterations is empirically chosen for the presented disaggregation problem as follows: 

\begin{equation}
a_{\mathrm{k}}(i) = 1 - \frac{0.9}{I} \cdot i
\label{eq:a_k}
\end{equation} 

\begin{equation}
a_{\mathrm{s}}(i) = 0.0002 - \frac{0.0198}{I} \cdot i
\label{eq:a_s}
\end{equation}

Due to the highly complex state space, we observe a high value of the inertia constant $a_{\mathrm{t}}$ to decelerate the convergence. Therefore, $a_{\mathrm{t}}$ is set to zero in this work as a simplification.\\
 
\subsubsection{Shaking}

When first initializing the PSO, the position matrices are set to zero for all particles at all entries in $S_{p}$. Events are then added in random 2\,\% of the entries. These 2\,\% are determined empirically based on the estimated number of events in the overall signal. To increase the robustness of the algorithm with respect to local optima, a number of epochs is defined, similar to \cite{yuriy2018}. In each epoch, the total signal passed is disaggregated with a selected number of iterations. At the beginning of each epoch, the particles are reinitialized. However, they start at the last known, globally best position. To this new starting position events are again randomly added at 2\,\% of the entries of the position matrices. This means that particles that were far away from a good explorative contribution in the state space can contribute to the optimization again. The multiple initialization of particles is hereafter referred to as \textit{Shaking}. For better convergence, the movements constants are varied in each epoch, as described above. For disaggregation, the maximum number of epochs is limited to 50 and the number of iterations per epoch is limited to 30. The PSO is initialized with ten particles. However, these hyperparameters depend on the selected speed parameters and the optimization problem itself. A higher number of epochs, iterations per epoch and number of particles increases the probability of finding better optima in the high dimensional state space. The values of the aforementioned hyperparameters are limited due to computational capacities and time.\\ 

\subsubsection{Dimension Reduction}
The main challenge in disaggregation is the high dimensionality due to the ever increasing temporal resolution and often a large number of individual device types. This also corresponds to the \textit{curse of dimensionality} formulated by Richard Bellman in \cite{bellman1961}. We present a method for dimensional reduction of the disaggregation problem. For this purpose the measured total signal is divided into sections with the length of 60 values which is equal to 60\,s according to the measuring frequency of 1\,Hz. In the following, these sections will be referred to as \textit{\mbox{frames}} and the method for dimension reduction will be referred to as \textit{framing}. The frame length can be varied with resprect to the computational resources. Using more particles could enable the optimization of longer frames. Nevertheless, framing results in a reduction of the number of possible combinations in the state change matrix of the individual particles, since $S$ is effectively reduced from the dimension $(86400 \times M)$ (for a whole day) to the dimension $(60 \times M)$. For each individual \mbox{frame} the disaggregation is performed including shaking and variation of the parameters.

For some device profiles $\tau_{i} > 60$\,s applies. 
Thus, the optimization result of a frame influences further frames, e.g. a device which is switched on in frame $f$ might show its dynamic behaviour through frame $f+1$ and reaches its steady state in frame $f+2$ or later frames. 
For this reason, we assume in the optimization of the next frame $f+1$ the effect of previous state changes to be fixed. If the PSO finds that this device type is switched off, the steady state power is substracted. 

More formally, let $P^0 = P - P(0)$ denote the power of the respective day with substracted always-on-component at $t = 0$. 
In the optimization of the first frame, we use the PSO to find $S_0$ such that $E^{[0,59)}(P^0, P_{\mathrm{S_0}})$ is minimized, i.e. only the entries of $S$ in the first 60 rows are modified such that the deviation of $P^0$ and $P_{\mathrm{S_0}}$ in the first frame is minimized.

In order to optimize the second frame, we set $P^1 = P^0 - P_{\mathrm{S_0}}$ and use the PSO to find $S_1$ (the first 60 rows are the same as $S_0$) by minimizing $E^{[60,119)}(P^1, P_\mathrm{S})$. Note that in this step only the rows 60 to 119 are being modified. 
After optimizing the second frame, the state change matrix $S_1$ contains non-zero entries in the first 120 rows, where the first 60 rows coincide with the first 60 rows of $S_0$. 

For the next frame we proceed with $P^2 = P^0 - P_{\mathrm{S_1}}$. 
The optimization is carried out for rows 120 to 179 as before. 
This procedure is repeated accordingly for all subsequent frames. 

\begin{figure}[!t]
\centering
\includegraphics[width=3.5in]{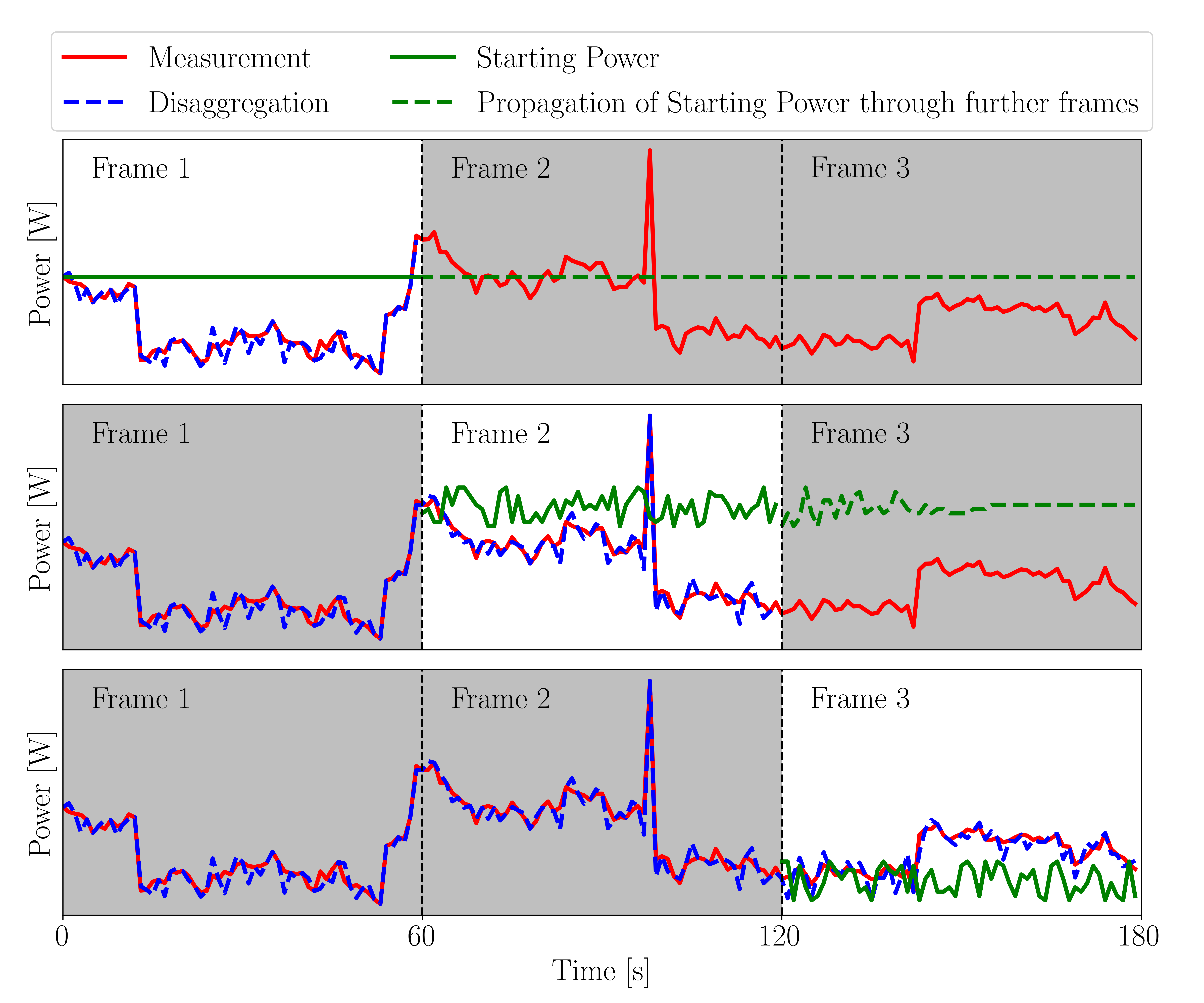}
\caption{Framing of the first three frames of one day. In the upper illustration, the disaggregation starts in frame 1 with no prior state changes. Thus, the starting power (green) is constant. In the middle illustation, the starting power results from the prior disaggregation of frame 1 and is no longer constant since the dynamic transient behavior from state changes in frame 1 affect frame 2. Respectively, the disaggregation works in the bottom illustration for frame 3 subsequently}
\label{fig:Framing}
\end{figure}

Figure~\ref{fig:Framing}, a graphical representation of the framing of the first three frames of one day, illustrates this procedure. 
$P(0)$ for the first frame and $P_{\mathrm{S_{f-1}}}$ for the $f$\textsuperscript{th} frame are referred to as starting power. The effects of $P_{\mathrm{S_f}}$ to the subsequent frames can be seen in Figure~\ref{fig:Framing} as propagation of the starting power. As described, the starting power for frame $f$ is set in frame $f-1$, it cannot be changed in frame $f$.\\

\subsubsection{Early Stopping}

A so-called \textit{Early-Stopping} criterion is used to reduce the required computational time. This has been implemented in the past in multiple variations by e.g. \cite{teixeira2008,zielinski2007}. In this work, we use a simple early stopping criterion. First, the algorithm runs for a certain minimum number of epochs for one \mbox{frame} to be disaggregated. If, after a selected number of epochs, the error of the currently known global best position does not improve, the disaggregation of this \mbox frame is stopped early and the disaggregation of the subsequent frame starts at the given starting position determined by the previous frames. Due to computational limits, the minimum number of epochs $e_{\mathrm{min}}$ is set to 5 and the maximum number of epochs in which the known global optimum $S_{\mathrm{g,best}}$ does not change $e_{\mathrm{nc}}$ is set to 3 in this work.\\

\subsection{Summarizing Presentation of the Algorithm}
In Algorithm~\ref{alg:PSO}, the whole procedure of particle swarm optimization for energy disaggregation is shown in pseudocode. The algorithm contains four nested loops which are reasoned by the division of the measured signal into multiple frames and the use of multiple optimization epochs per frame. Therefore, the computational time is increasing linearly with a higher number of frames, epochs, particles or iterations per epoch respectively. 

\begin{algorithm}[!t]
\caption{Disaggregation via adapted PSO}\label{alg:PSO}
\begin{algorithmic}[1]

\For{$f = 1 \dots \text{Number of Frames}$}: 
	\State $S_{\mathrm{g,best}} = 0$
	\For {$e = 1 \dots \text{Number of epochs}$}: 
	\State Initialize particle around the current global best position $S_{\mathrm{g,best}}$
	\For{iterations $i = 1 \dots I$}:  
	\For{particle $p = 1\dots \text{Number of particles}$}: 

	\State Velocity update according to Eq.~\ref{eq:vel_update}
	\State Position update according to Eq.~\ref{eq:pos_update}
	\State Reconstruct power according to Eq.~\ref{eq:reconstruction}
	\State Calculate error $E$ according to Eq.~\ref{eq:error_disaggregation}
	
	\If{$E_{S_{f,i-1,p}} < E_{S_{\mathrm{g,best}}}$}: 
	\State $S_{\mathrm{g,best}} \gets S_{f,i-1,p}$
	\EndIf	
	\EndFor
	\EndFor
	\State \# Check Early-Stopping criterion: 
	\If {$E_{S_{\mathrm{g,best}}}(e) = E_{\mathrm{{S_{\mathrm{g,best}}}}}(e-e_{\mathrm{nc}}) $ and $e > e_{\mathrm{min}}$}: 
	\State break 
	\EndIf
	\EndFor
	\EndFor 
	
\end{algorithmic}
\end{algorithm}

\section{Testing Of The Developed Method}

In order to test the PSO algorithms and the presented adaptions for energy disaggregation, two datasets of industrial power data were used which are described in the first of the following sections. Since there the data is not labelled, evaluation is conducted in the power domain by different error measures. They are stated in the subsequent section. Thereafter, the results of one dataset are presented extensively and followed by an examination of the transferability of the developed algorithm by testing it with the second dataset. This section will be concluded with a discussion of the results.

\subsection{Data Description}
For developing and testing the improvements of the PSO presented in this work, power data of two different measuring points in a multi tenant commercial building is used. Both datasets represent industrial power demands similar to shop floors. The datasets each consist of six features: Three phases of active and reactive power. Therefore, the power at time $t$ is represented by $P(t) \in \mathbb{R}^6$ as already assumed above. Accordingly, each power dataset can be described by $P \in \mathbb{R}^{T_{\mathrm{tot}}\times 6}$ with $T_{\mathrm{tot}}$ being the total length of the dataset. This power matrix contains six columns $P_0 \dots P_2$, where $P_0 \dots P_2$ describe active power and $P_3 \dots P_5$ describe reactive power. 

The main testing of the algorithm happens by means of Dataset 1 and the transferability of the algorithm is tested on Dataset 2. In Table~\ref{tab:data_analysis} key figures for both measuring points are shown to describe both datasets. The granularity of both datasets is 1\,s. No other data or prior knowledge other than the measured power and the device profiles is present or used. Dataset 1 contains measurements from December 1\textsuperscript{st}, 2018 until April 30\textsuperscript{th}, 2019. From Dataset 2, we use one day of data: January 14\textsuperscript{th}, 2019. On average there are 0.0023\,\% of datapoints missing. Gaps are filled by the last known value. The measuring power analyzer is of the type UMG 604 PRO from Janitza Electronics. According to the manufacturer the measuring deviation is up to 0.4\,\% and will be neglected in this work \cite{Janitza}. Due to the industrial occupation of the building, there is much regularity within the power measurements regarding repetetive patterns on working days and a low power demand on weekends. 

\begin{table}[!t]
\renewcommand{\arraystretch}{1.3}
\caption{Data Analysis of Dataset 1 and 2}
\label{tab:data_analysis}
\centering
\begin{tabular}{|c|c|c|}
\hline
& Dataset 1 & Dataset 2\\
\hline
Min [kW] & 2.26 & 0.97 \\
Max [kW]& 98.95 & 17.13 \\
Mean Energy per Day [kWh] & 534.51 & 74.14\\
Mean Power [kW]& 22.27 & 3.09\\
\hline
\end{tabular}
\end{table}

In the power data of Dataset 1, 52 single device profiles are extracted by statistical methods similar to \cite{Goncalves2011} as well as 27 single device profiles for Dataset 2. Nevertheless, the way of determining profiles of individual devices is of minor importance in this work as they are assumed to be prior knowledge. They can be extracted by various ways from the aggregate power data or directly measured.

\subsection{Error Measures}

In order to quantify the results of the method, various error characteristics are used, which are briefly described in this section. Since there is only the power data in six dimensions available for this work, all error calculations take place in the power domain.  Therefore, the disaggregated and reconstructed power $P_{\mathrm{S}}$ is compared to the measured power $P_{\mathrm{meas}}$. This is a significant difference to most other publications where the knowledge about the correct state of all single appliances is often available and used from datasets like \cite{kolter2011redd,kelly2015uk}. Therefore, often used error measures in literature are the accuracy or \mbox{$f$-measure} \cite{Faustine2017} which are not applicable in this work. In the following equations, all utilized error measures are outlined. It should be noted here that the in the following presented error characteristics are calculated for sum of the three active power phases $P_{\mathrm{tot}} = P_0 + P_1 + P_2$. Since this represents the total power actually demanded, an error calculation of it is sufficiently meaningful in order to describe the quality of reconstruction. Firstly, the root mean squared error (RMSE) is used as a default error measure for comparing data with respect to the absolute values within the data. The same applies for the mean absolute error (MAE) but due to the missing quadrature of differences like for the RMSE, the MAE is less sensitive to outliers. Comparing those two measures allows conclusions to be drawn about the frequency or severeness of outliers.

\begin{equation}
 \operatorname{RMSE}(P_{\mathrm{meas}},P_{\mathrm{S}}) = \sqrt{\frac{\sum_{t=0}^{T}(P_{\mathrm{meas}}(t)-P_{\mathrm{S}}(t))^2}{T}}
 \label{eq:RMSE}  
\end{equation}

\begin{equation}
  \operatorname{MAE}(P_{\mathrm{meas}},P_{\mathrm{S}}) = \frac{\sum_{t=0}^{T}|P_{\mathrm{meas}}(t)-P_{\mathrm{S}}(t)|}{T}  
 \label{eq:MAE}  
\end{equation}

As an error measure independent of the absolute power values, the mean absolute percentage error (MAPE) is introduced. The MAPE of different datasets is comparable which is important for later evaluating the transferability of the developed method.  

\begin{equation}
  \operatorname{MAPE}(P_{\mathrm{meas}},P_{\mathrm{S}}) = \frac{1}{T} \sum_{t=0}^{T}\frac{|P_{\mathrm{meas}}(t)-P_{\mathrm{S}}(t)|}{P_{\mathrm{meas}}(t)}  
 \label{eq:MAPE}  
\end{equation}

Lastly, in order to be able to evaluate the consumed energy of the measured power and the reconstructed power, a final error measure is introduced. It calculates the percentage difference between the energy actually consumed and the energy consumed of the reconstructed power over a specific time. It is particularly important for applications in the field of energy management because it indicates the quality of reconstruction over time and regarding also the financial aspect of cocts of electricty.    

\begin{equation}
 \operatorname{Energy}_{\mathrm{E}}(P_{\mathrm{meas}},P_{\mathrm{S}}) = \frac{|\sum_t P_{\mathrm{meas}}(t) - \sum_t P_{\mathrm{S}}(t)|}{\sum_t P_{\mathrm{meas}}(t)}
 \label{eq:MAPE}  
\end{equation}

\subsection{Results}

\begin{figure}[!t]
\centering
\includegraphics[width=3.5in]{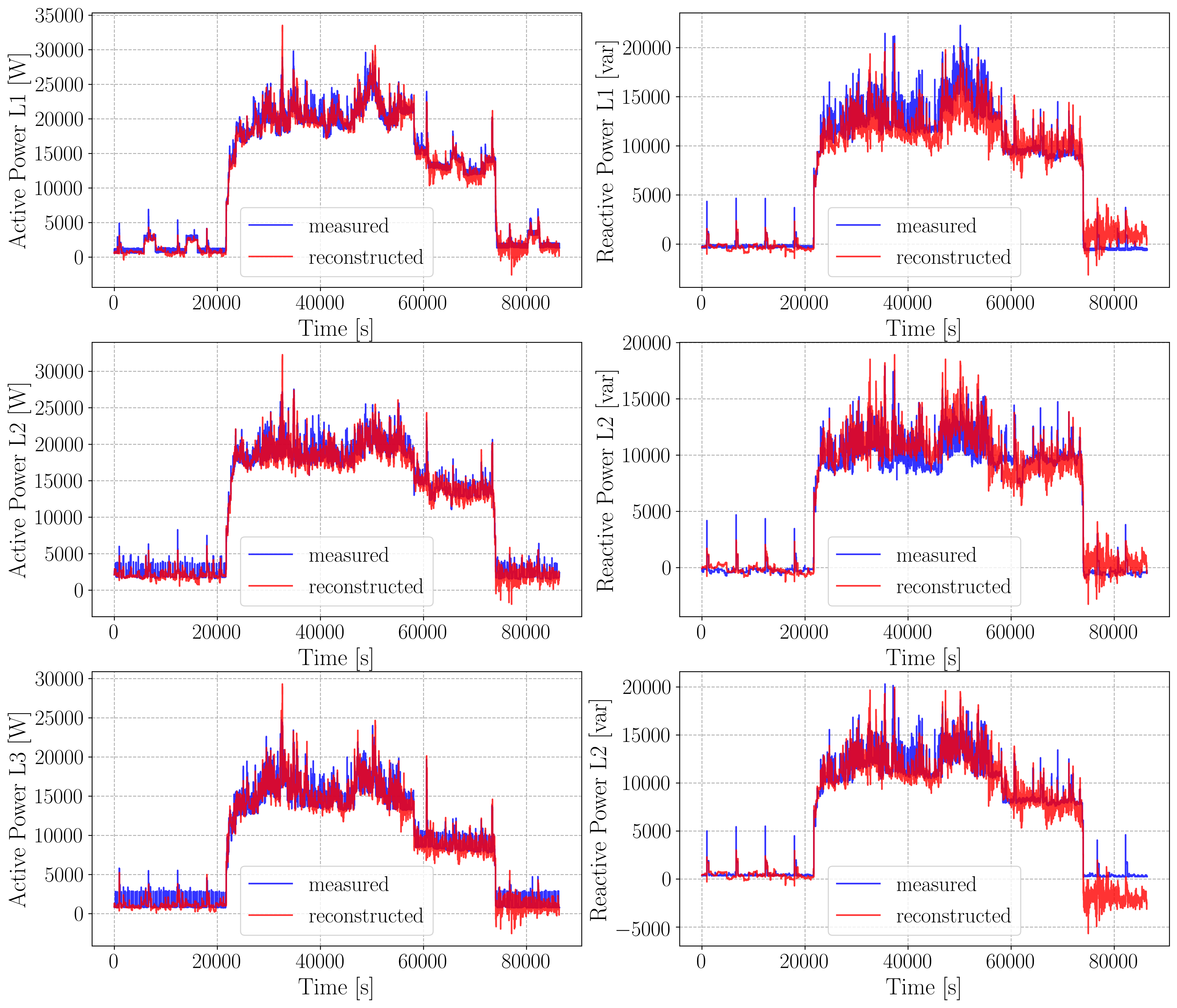}
\caption{Representation of disaggregated and reconstructed $P_0 \dots P_5$ of $P \in \mathbb{R}^{T \times 6}$ of December 4\textsuperscript{th}, 2018. The respective measurement is shown in blue and the  power disaggregated and reconstructed by the Algorithm~\ref{alg:PSO} in red. $L_1 \dots L_3$ denote the three phase.} 
\label{fig:PSO_Bsp} 
\end{figure}

\begin{figure}[!t]
\centering
\includegraphics[width=3.5in]{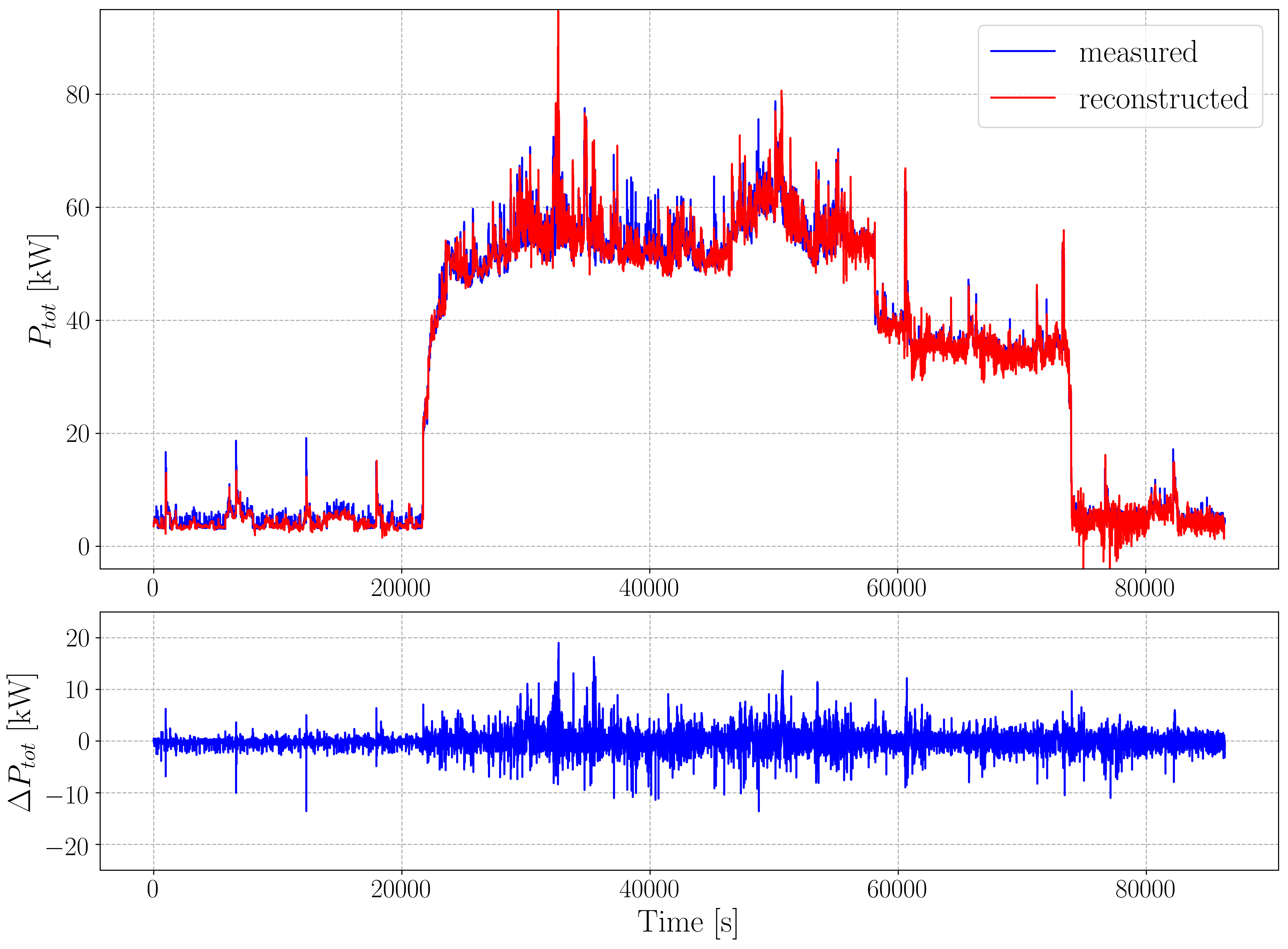}
\caption{Upper illustration: representation of disaggregated and reconstructed $P_{\mathrm{tot}} = P_0 + P_1 + P_2$ of December 4\textsuperscript{th}, 2018. Bottom illustration: Deviation between measured and  reconstructed power after disaggregation of the respective day.} 
\label{fig:PSO_Bsp_with_error} 
\end{figure}

In order to test the proposed method Dataset 1, as described above is used. The algorithm is applied to one day of data from December 2018 and the working days of March 2019. Weekends are of minor interest, since there are significantly less events due to a lack of working employees at this times. This procedure investigates the reproducability of the quality of results within a period of time and between two periods. The disaggregation of one full month is undertaken one day at a time. Thereafter, for each day the power is reconstructed in six dimensions and the preseneted error measures are calculated for the sum of all active power phases. The proposed error measures are calculated for each day individually. Of these daily error values, the mean and standard deviation are presented in Table~\ref{tab:PSO_errors}. Figure~\ref{fig:PSO_Bsp} shows one day of disaggregated and reconstructed power in comparison to the measured power. The great similarity of the two power curves is clearly visible in all six phases. In particular, short time peaks and fast, sharp rises in the signal are detected by the PSO, although the absolute height of the peaks is sometimes underestimated. It can also be seen that the reconstructed signal seems to be noisy at the end of the working time. This can be attributed to overlays of device profiles that are still on after the end of working time. In Figure~\ref{fig:PSO_Bsp_with_error}, the measured and reconstructed total active power $P_{\mathrm{tot}}$ is shown as this quantity is of great importance for energy management. Beneeth that in Figure~\ref{fig:PSO_Bsp_with_error} the absolute deviation between both power curves is shown in order to visualize the differences. It is visible, that no constant offset is present between measurement and reconstruction as well as there is lower devation at night than during working time. The most deviations are of short duration, therefore no error propagation can be seen.

The error values in Table~\ref{tab:PSO_errors} show repeatable good results even on large amounts of data for the proposed algorithm. The results of Dataset 1 are reproducible within comparatively small error deviations, since the standard deviation from the mean value within the month of data is on average about 10\% of the absolute error values. Furthermore, all error characteristics of the single day and the mean values of the month under consideration are very similar, although there is a time span of three months between the data. The deviations in the energy consumed per day are on average less than one percent for all data considered. Since the RMSE is 70\,\% larger than the MAE, the conclusion can be drawn that there are some outliers between the reconstructed and the measured power. As can be seen in the Figure~\ref{fig:PSO_Bsp_with_error}, these outliers occur mainly at the high peaks in the load curve during working time.

\begin{table}[!t]
\renewcommand{\arraystretch}{1.3}
\centering
\caption{Error characteristics between the sum of active power measured and reconstructed after disaggregation for different periods of time of Dataset 1. The mean and standard deviation of the error characteristics for March 2019 are given with respect to the daily disaggregation error}
\begin{tabular}{|c|c|c|}
\hline
  & 1 Day  & 1 Month \\
  &(December 1\textsuperscript{st}, 2018) & (March 1\textsuperscript{st}, 2019\\
  && - March 31\textsuperscript{th}, 2019)\\
\hline
RMSE [W] 	& 1542 	& 1565 $\pm$ 150	\\
MAE [W] 	& 926  	& 921 $\pm$ 85 	\\
MAPE [\%]	& 6.18 		& 6.04 $\pm$ 1.51		\\
$\operatorname{Energy}_{\mathrm{E}}$\,[\%]	& 0.716 	& 0.897 $\pm$ 0.156		\\
RMSE / Mean Power\,[\%] &6.92& 7.02 $\pm$ 0.67\\  

\hline
\end{tabular}
\label{tab:PSO_errors}
\end{table}

\begin{figure}[!t]
\centering
\includegraphics[width=3.5in]{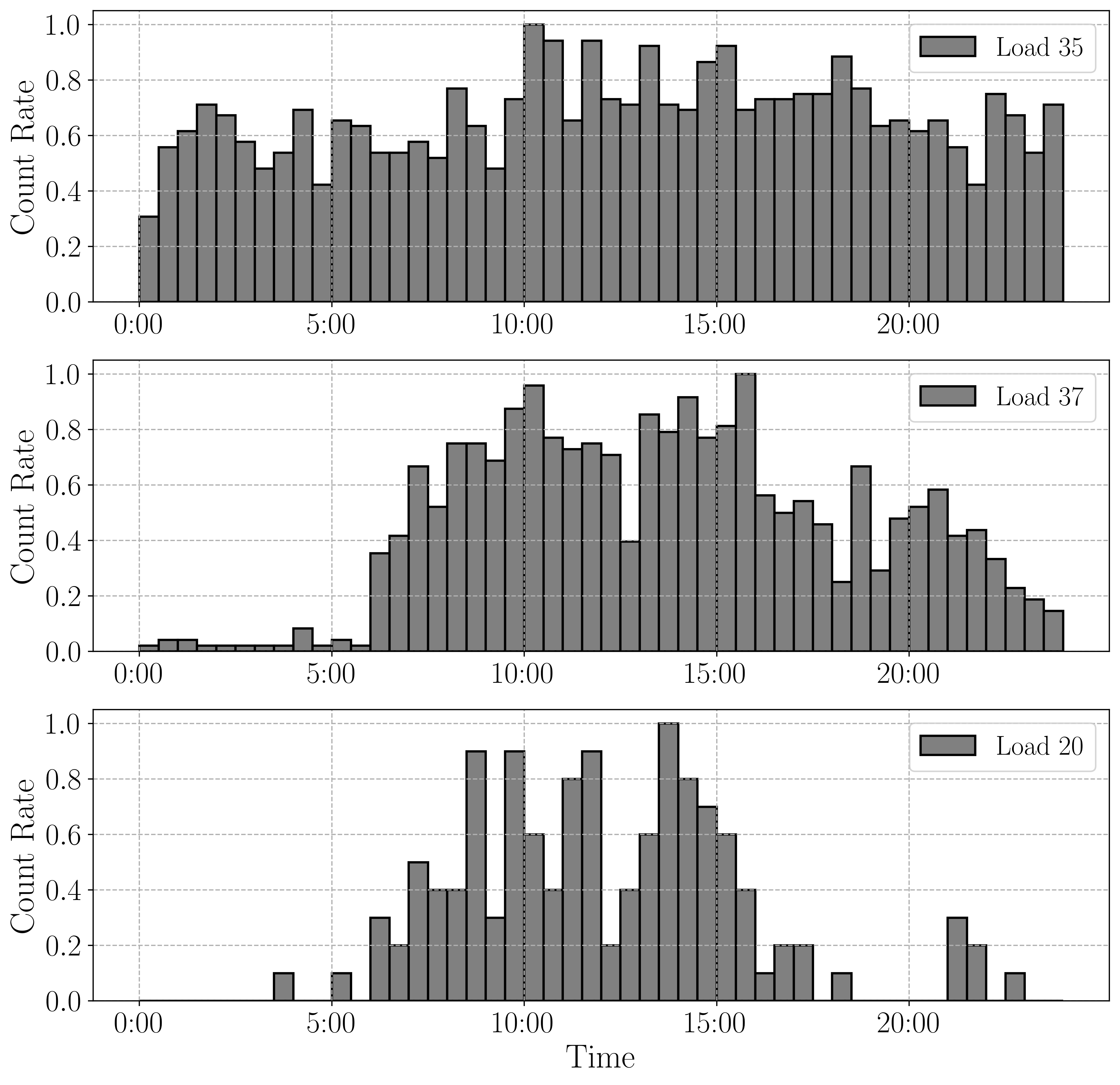}
\caption{Exemplary representation of distributions of ON events of three individual devices.} 
\label{fig:ON_events} 
\end{figure}

For further illustration of the results of disaggregation, the daily distributions of ON events for exemplary device profiles are calculated. They quantify in which time window of the day an individual device has a certain frequency of a state change to the state ON. The time window is limited to 30\,min. An individual device can be switched on or off several times within one time window. Therefore, the values of every time window are divided by the maximum of ON-events within one window to improve the comparability. For clearer insights, the displayed histograms in Figure~\ref{fig:ON_events} are calculated based on the results from the working days of March 2019. Three exemplary distributions are shown in Figure~\ref{fig:ON_events} for different individual devices. Therein, three different types of distributions can be observed, each with specific characteristics. On the one hand, a working time independent behavior can be seen as shown for Device 35 in Figure~\ref{fig:ON_events}. Individual devices with this behavior are switched on evenly distributed throughout the day. On the other hand, a working time dependent behaviour can be observed in the distribution shown for Device 37 and 20 in Figure~\ref{fig:ON_events}. 

\subsection{Transferability}

\begin{figure}[!t]
\centering
\includegraphics[width=3.5in]{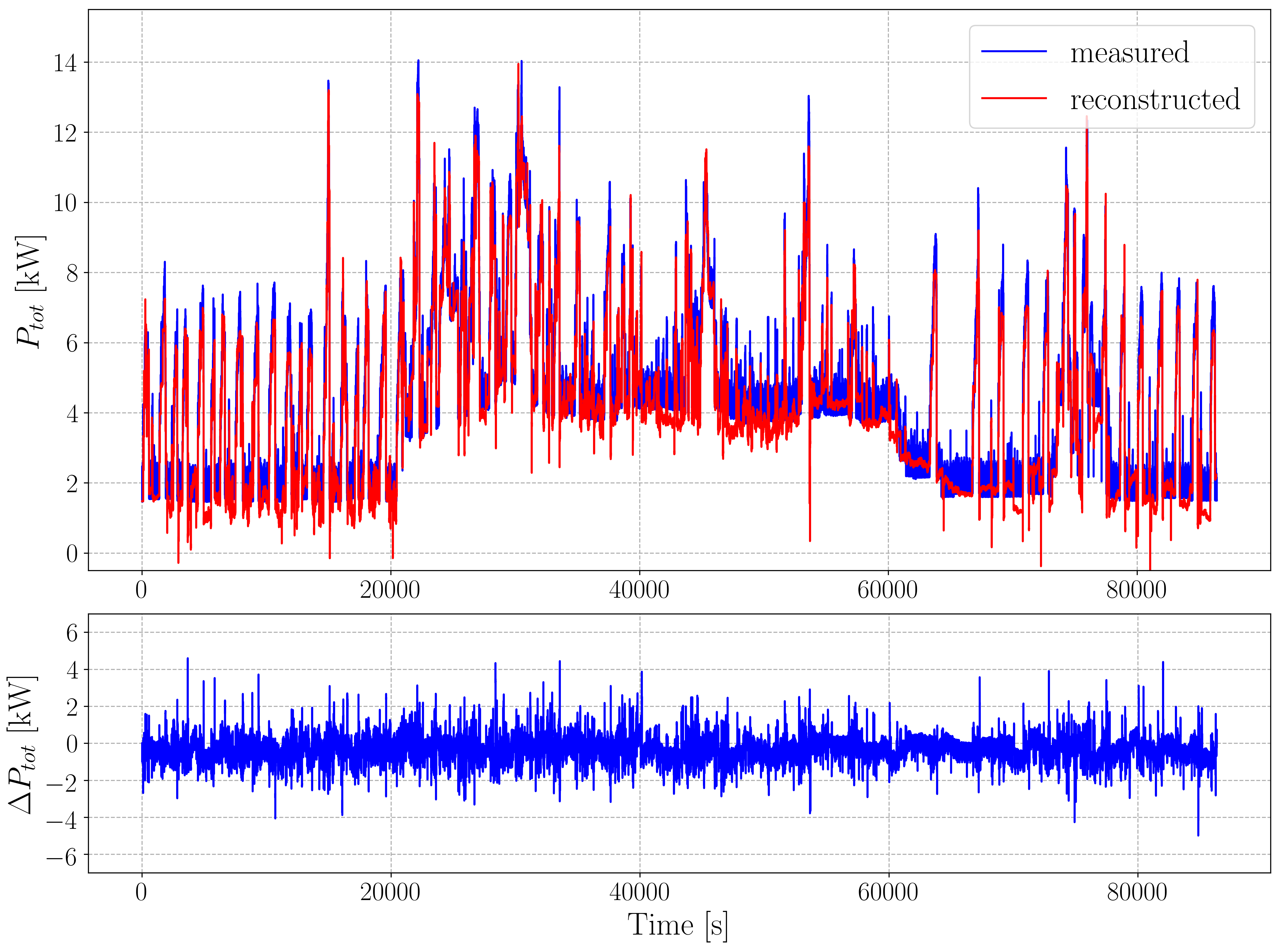}
\caption{Upper illustration: Reconstructed and measured power (sum of active phases) for one day (January 14\textsuperscript{th}, 2019) of Dataset 2. Bottom illustration: Absolute deviation between reconstructed and measured power.} 
\label{fig:PSO_Bsp_EMDAQ5} 
\end{figure}

To investigate the transferability of the proposed method, one day of Dataset 2 is disaggregated analogously. All parameter choices of the PSO are kept the same. Figure~\ref{fig:PSO_Bsp_EMDAQ5} shows one day of data (January\,14\textsuperscript{th}\,,2019), where the sum of three phases in active power of the measured and reconstructed power is shown. It can be seen that the proposed disaggregation method is working well. Especially large peaks are accordingly reconstructed. However, there are more deviations at low power values than for Dataset 1. To quantify the deviations between measured and reconstructed power, the values of the error measures introduced above are shown in Table~\ref{tab:PSO_Fehler_EMDAQ5}. It can be seen that the relative error in the consumed energy is similarly low as for the results of Dataset 1. The RMSE and MAE are significantly smaller, but this is due to the smaller absolute power values. Dividing the RMSE by the mean power value shows an 7 percentage points higher error than for Dataset 1. The MAPE is approximately twice as large as for the results of Dataset 1. 

\begin{table}[!t]
\renewcommand{\arraystretch}{1.3}
\centering
\caption{Multiple error characteristics between sum of active power phases reconstructed and measured of dataset 2 for one day (January 14\textsuperscript{th}, 2019)}
\begin{tabular}{|c|c|}
\hline
Error Measure& Value\\
\hline
RMSE [W] 	& 568.1 \\
MAE [W] 	& 422.1  \\
MAPE [\%]	& 13.01	  \\
$\operatorname{Energy}_{\mathrm{E}}$[\%]	& -0.88 \\
RMSE / Mean Power\,[\%] & 13.72\\

\hline
\end{tabular}
\label{tab:PSO_Fehler_EMDAQ5}
\end{table}

\subsection{Discussion}

Given the results from the previous sections a comparison to other publications is of great interest. This is proving difficult because of a use of different error measures and the lack of labelled power data in this work. But in \cite{Kelly2015} the relative error in total energy is calculated for the disaggregation of household power data using different neural networks and selected appliances. The best error values are of 12\%, which is significantly higher than the proposed error values in consumed energy of less than 1\%. This demonstrates the optimization of the PSO in the power domain useful and powerful to disaggregate and reconstruct the aggregate power signal with given device profiles. The displayed distributions of ON-events in Figure~\ref{fig:ON_events} allow for interpretations of the behavior of the individual devices after disaggregation. It can be seen that the ON-events are not randomly distributed by the PSO, but follow certain patterns. These patterns in the distributions can be illustrated above all by a division into individual devices that depend on working time and individual devices that belong to the base load. The distribution for Device 37 in Figure~\ref{fig:ON_events} shows a relativley high ON-event frequency after the end of working time. This could correspond to a mixture of the patterns of Device 35 and 20 in Figure~\ref{fig:ON_events} and could either result from a partially incorrect disaggregation at the end of the day or may be justified by dependecies on other quantities but time. Figure~\ref{fig:PSO_Bsp} shows noise in the reconstructed power at the end of the day, which could be explained in this way. Testing the developed algorithm with the same parameter choices for a different dataset shows comparably good results especially in the error of consumed energy. Other error measures like MAPE are higher for Dataset 2 than fpr Dataset 1 but with adaptions of the chosen parameters of the PSO, further improvements in the error characteristics can be expected. Adaptive methods for the parameters of PSO for energy disaggregation have to be examined in the future.  

In this work, we assume a standard granularity of 1\,s and the presence of power data in active and reactive power in all three phases. This results in a very high-dimensional state space of the PSO but allows for distinct differentiation of events in the aggregate power signal. Due to the high measuring frequency, effects from simultanious events are decreased and could be minimized with even higher measuring frequencies. With a lower measuring frequency the distinction of events and the correct assignment of device profiles to certain points in time becomes more difficult. The limits of granularity could be estimated according to the number of events in the aggregate power signal and could be further investigated in the future. 

The PSO procedure in this work is not constrained by any form of function. That allows for inaccuracies in disaggregation as for instance more OFF-events than ON-events for a device profile. These inaccuracies could be solved by the implementation of a constraining function. This could potentially increase the computational time since additional steps would take place in the algorithm. This approach remains to be examined. But with more distinguishing characteristics for the single devices and with higher the measuring frequency these inaccuracies could be reduced without such a constraining function. 

The assumptions made for the device profiles (transient or dynamic state and stable state) are an approximation of the real behavior. Depending on the device profiles, which build the aggregate signal, this could lead to minor or major problems in the disaggregation. In this work, we did not see major error sources for two independent datasets of industrial power consumption. 

Nevertheless, the proposed approach provides several advantages over other disaggregation methods. Firstly, there is no prior knowledge besides the device profiles required and no labelled datasets are necessary for disaggregation. The needed device profiles have to be extracted or measured beforehand. Since there is no complex model building as in most other publications \cite{Kelly2015}, no fitting or training of many parameters is required as for the most machine learning approaches. Secondly, only three parameters (interia and movement constants) need to be adapted to the problem. Other parameters of the PSO, as number of particles and epochs, can be adapted to the data optionally. This would open up possibilities of applications to different kinds of buildings and would not even be restricted to energy disaggregation in buildings but also higher levels of the electricity grid like districts. 

Due to the daily disaggregation, long term erros can not propagate. A higher number of particles than applied in this paper likely would yield in a faster convergence and the probability of finding the global optimum would increase \cite{lee2017}. But the compuational time would increase with more particles, more epochs, a higher granularity or more device profiles. Therefore, the possibility of parallel processing should be investigated in the future. Parallelisation has been proven to be applicable to PSO in multiple publications in the past and could be useful in order to reduce the computational time \cite{chang2005,kim2007}. This could enable several oportunities of online disaggregation.

\section{Conclusions and Outlook}

In this paper, we present a fully unsupervised disaggregation method using particle swarm optimization and adapting the classic metaheuristic to the stated disaggregation problem. For this paper we used power data of two industrial consumers to apply and validate the proposed method. All evaluation takes place in the power domain since the used datasets are unlabelled ragarding the state of single devices. In general, labelled datasets are rarely available especially for industrial and commercial buildings. Thus, the examination of methods for energy disaggregation which not rely on labelled data is very important. Aggregating the power after disaggregation according to Equation~\ref{eq:aggregate} shows a very accurate reconstruction of the measured power curve of Dataset 1. Since other research mostly works with labelled datasets and different error measures like for instance \cite{lu2017} a comparison is difficult. In order to improve the PSO several adaptions are of interest. Especially, parallel processing could speed up the disaggregation process and adaptive methods for choosing the hyperparameters could increase the performance and transferability.   
With the proposed disaggregation procedure a new level of data is created which represents additional knowledge to the measured aggregate power consumption. This metalevel of data could be used for various applications like analysis or detection of behavioural changes or the detection of new appliances and devices. Especially the training of powerful machine learning algorithms like artificial neural networks could provide opportunities like for instance a prediction of state changes in order to improve power forecasts.

\section*{Acknowledgment}

The authors acknowledge the financial support of the Federal Ministry for Economic Affairs and Energy of the Federal Republic of Germany for the project \textit{EG2050: EMGIMO: Neue Energieversorgungskonzepte für Mehr-Mieter-Gewerbeimmobilien (03EGB0004G and 03EGB0004A)}. For more details, visit www.emgimo.eu. The presented study was conducted as part of this project.

\ifCLASSOPTIONcaptionsoff
  \newpage
\fi

\begin{IEEEbiography}
[{\includegraphics[width=1in,height=1.25in,clip]{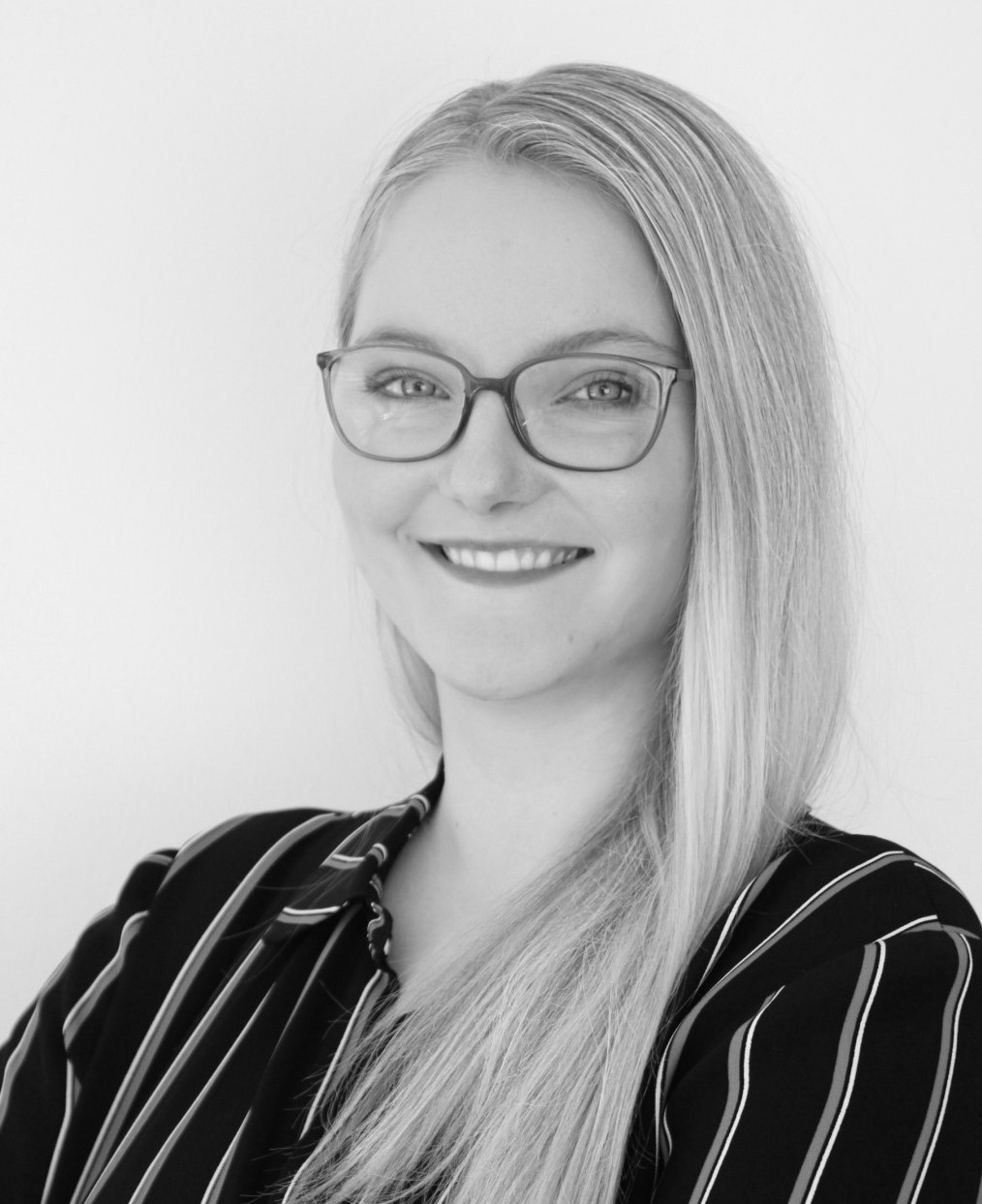}}]
{Karoline Brucke}
received the MSc degree in physics from the Carl-von-Ossietzky University Oldenburg, Germany, in 2020. During her master thesis in 2019 she specialized on machine learning based energy management including energy disaggregation and load forecasting.  Since 2015 she is working at the DLR Institute of Networked Energy Systems, since 2018 in the group of energy management.  Her research interests include non-intrusive load monitoring and its applications for i.e. energy forecasting based on machine learning and metaheuristic optimization.
\end{IEEEbiography}

\begin{IEEEbiographynophoto}{Stefan Arens}
received the BSc in process engineering and energy technology from the Bremerhaven University of Applied Sciences in 2014 and the M.Sc. degree in engineering physics with a specialization in renewable energy from the Carl von Ossietzky University, Oldenburg, in 2018. Currently, he is pursuing his Ph.D in the Energy Systems Technology department at the DLR Institute of Networked Energy Systems in Oldenburg, Germany. His research interests include smart energy management systems, energy forecasting, and application scenarios of energy management systems.
\end{IEEEbiographynophoto}

\begin{IEEEbiography}
[{\includegraphics[width=1in,height=1.25in,clip]{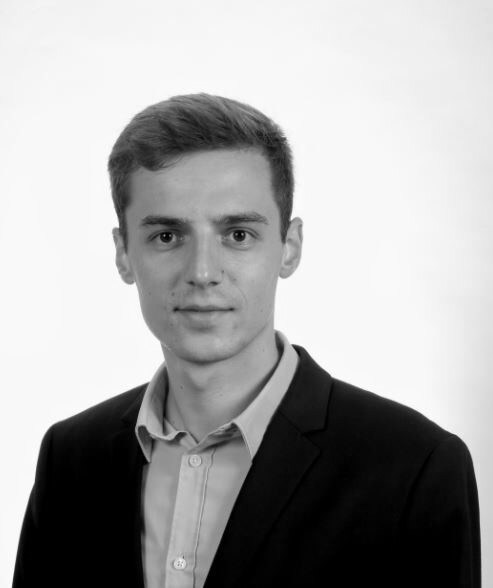}}]
{Jan-Simon Telle}
received his MSc degree in sustainable energy systems from Otto-von-Guericke Universität Magdeburg, Germany, in 2017. After his MSc graduation he is working as research associate and project manager in the field of energy management research at the DLR Institute of Networked Energy Systems in Oldenburg, Germany. His research interests include forecast and machine learning based load management solutions at building level and smart operating strategies for integrated energy systems.
\end{IEEEbiography}

\begin{IEEEbiography}
[{\includegraphics[width=1in,height=1.25in,clip]{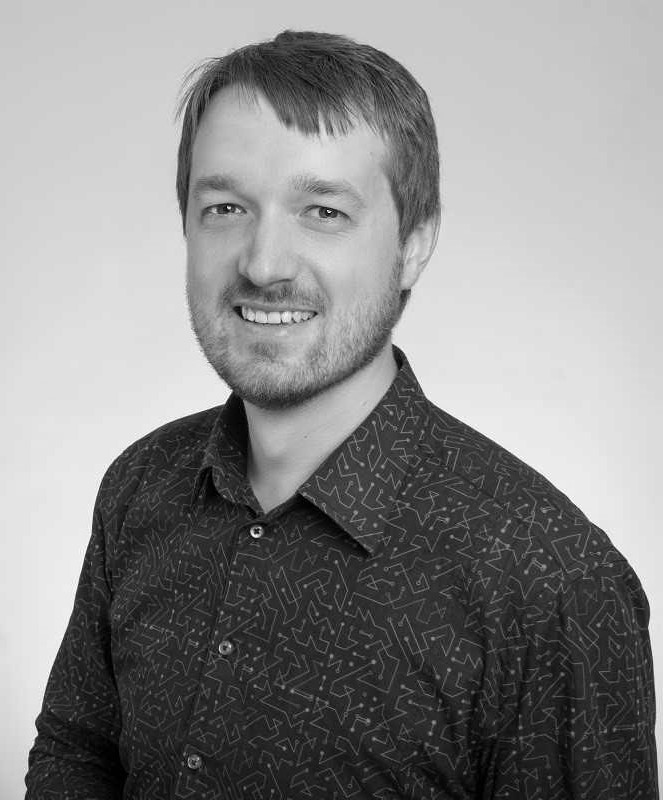}}]
{Sunke Schlüters}
received the PhD degree in pure mathematics from Carl von Ossietzky University Oldenburg, Germany, in 2015 and switched to the field of energy management research in 2018. He is currently working as a researcher and project manager on energy management strategies and algorithms with the DLR Institute of Networked Energy Systems. His research interests include machine learning based self-optimizing energy management systems.
\end{IEEEbiography}

\begin{IEEEbiography}
[{\includegraphics[width=1in,height=1.25in,clip]{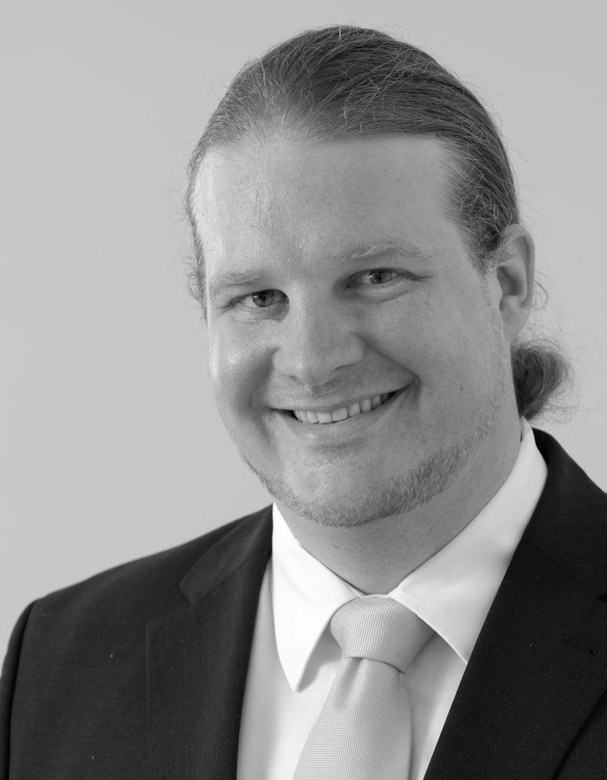}}]
{Benedikt Hanke}
received the PhD degree in engineering from the Albert-Ludwigs-University Freiburg im Breisgau in 2010, researching the dielectric constant of polymer ceramic composites. He worked in the field of material research for silicon thin film photovoltaics for several years. Currently he is a team leader at the DLR-Institute of Networked Energy Systems heading the energy management group. His research interests include low maintainance, low cost energy management systems based on machine learning approaches as well as methods for minimal energy system design by optimization of operational strategies.
\end{IEEEbiography}

\begin{IEEEbiography}
[{\includegraphics[width=1in,height=1.25in,clip]{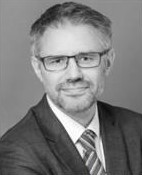}}]
{Karsten von Maydell}
received the MSc degree in physics from the University of Oldenburg, Germany in 2000 and the PhD degree in physics from the University of Marburg, Germany, in 2003. From 2000 to 2006, he worked as a graduate research assistant and postdoctoral researcher and project manager at the Helmholtz-Zentrum Berlin. From 2006 to 2007, he worked as a project manager R\& D at Q-Cells AG Thalheim, Germany and from 2007 to 2008, he was group leader at the Energy and Semiconductor Research laboratory at the University of Oldenburg. He was head of division Photovoltaic from 2008 to 2014 at the NEXT ENERGY research institute in Oldenburg, Germany. Since 2017, he is the head of Energy Systems Technology department at the DLR Institute of Networked Energy Systems. His research interests include the design of energy systems, smart energy management for grid connected and off-grid connected systems, integration of flexibilities in energy systems and robust operation of power grids. 
\end{IEEEbiography}

\begin{IEEEbiography}
[{\includegraphics[width=1in,height=1.25in,clip]{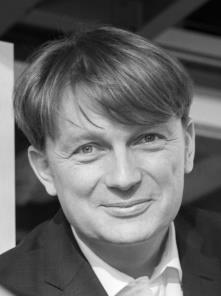}}]
{Carsten Agert}
studied Physics at the University of Marburg, Germany, and the University of Canterbury, UK. As a scholar of the Studienstiftung he completed his Ph.D. in Physics in 2001, based on his research work at the Fraunhofer Institute for Solar Energy Systems in Freiburg, Germany, and at the University of Oxford, UK. In 2001 he worked as a Postdoctoral Researcher in Pretoria, South Africa. From 2002 to 2005, he was a Research Associate with the German Advisory Council on Global Change WBGU. Subsequently, he was the head of the Fuel Cell Systems Research Group at the Fraunhofer Institute for Solar Energy Systems, from 2005 to 2008. Since 2008, he is a full professor at the University of Oldenburg and director of the DLR Institute of Networked Energy Systems, until 2017 known as NEXT ENERGY institute. His research interests include materials and device development for energy converters, research into electrical energy technologies and systems as well as energy systems analysis. 
\end{IEEEbiography}

\end{document}